\documentclass[a4paper,10pt]{article}
\usepackage{aipm}
\usepackage{graphicx}
\usepackage{subfig}
\usepackage{algorithm}
\usepackage{algorithmic}
\usepackage{url}
\usepackage{perpage} 
\usepackage{multirow}
\MakePerPage{footnote}

\affiliation{
\dag University of Macedonia, Department of Applied Informatics, 54006, Thessaloniki, Greece \\
{\ \{stoug, gevan\} (at) uom.gr} \\
\ddag Alexander TEI of Thessaloniki, Department of Informatics, 57400, Sindos, Greece \\
{\ dad (at) it.teithe.gr}
}
\begin{document}

\title{An Extensive Experimental Study on the Cluster-based Reference Set Reduction for speeding-up the $k$-NN Classifier}
\author{Stefanos Ougiaroglou$^{\dag}$\thanks{\hspace{3 mm}Stefanos Ougiaroglou is supported by a scholarship from the State Scholarship Foundation of Greece (I.K.Y.).} , Georgios Evangelidis$^{\dag}$ and Dimitris A. Dervos$^{\ddag }$}
\maketitle

\abstract
The $k$-Nearest Neighbor ($k$-NN) classification algorithm is one of the most widely-used lazy classifiers because of its simplicity and ease of implementation. It is considered to be an effective classifier and has many applications. However, its major drawback is that when sequential search is used to find the neighbors, it involves high computational cost. Speeding-up $k$-NN search is still an active research field. Hwang and Cho have recently proposed an adaptive cluster-based method for fast Nearest Neighbor searching. The effectiveness of this method is based on the adjustment of three parameters. However, the authors evaluated their method by setting specific parameter values and using only one dataset. In this paper, an extensive experimental study of this method is presented. The results, which are based on five real life datasets, illustrate that if the parameters of the method are carefully defined, one can achieve even better classification performance.
\endabstract


\doi
It would be provided by publication house.
\enddoi

\section{INTRODUCTION}

The data mining algorithms that assign new data items into one of a given number of categories (or classes) are called classifiers (Han and Kamber, 2000). Classifiers can be evaluated by two major criteria: classification accuracy and computational cost. $k$-NN is an extensively used and effective lazy classifier (Dasarathy, 1991). It works by searching the training data in order to find the $k$ nearest neighbors to the unclassified item $x$ according to a distance metric. Then, $x$ is assigned into the most common class among the classes of the $k$ nearest neighbors. Ties are resolved either by choosing the class of the one nearest neighbor or randomly. This work adopts the first approach. 

However, the $k$-NN classifier has the major disadvantage of high computational cost as a consequence of the computations needed to estimate all distances between a new, unclassified, item and the training data. Thus, as the size of the training set becomes larger, the computational cost increases linearly. Many researchers have focused on the reduction of the $k$-NN computational cost and therefore several speed-up methods have been proposed. These methods are mainly based on either indexing (Samet, 2005; Zezula et al, 2006) or data reduction techniques (Wilson and Martinez, 2000; Lozano, 2007). Additionally to these methods, recent research proposed cluster-based approaches for speeding-up the $k$-NN classifier, such as, the Cluster-based Trees (Zhang and Srihari, 2004), the Representative-based Supervised Clustering Algorithms (Eick et al, 2004), and, the Reference Set Reduction method through $k$-means clustering (Hwang and Cho, 2007). This work focuses on the latter approach.

The Reference Set Reduction Method is an adaptive approach which provides three parameters. Its effectiveness depends on the adjustment of these parameters. Hwang and Cho presented experimental results obtained by specific parameter values and based on only one dataset. Moreover, they did not use the well known Euclidean distance as the distance metric. These observations constitute the motivation of our work. Thus, the contribution of this paper is an extensive experimental study on this method. It includes experiments on five real life datasets using different parameter values. We also use as a metric the Euclidean distance.

The rest of this paper is organized as follows. Section II considers in detail the Reference Set Reduction method through $k$-means clustering and discusses the adaptive schema that it provides. In Section III, we present an extensive experimental study based on five real life datasets. The paper concludes in Section IV.

\begin{algorithm*}[t]
\textbf{Input:} \textit{k, L, D}
\caption{Reference Set Reduction through $k$-means clustering}
\label{rsrm}
\begin{algorithmic}[1]
   \STATE \COMMENT{Preprocessing procedure}   
      \STATE Use the first $\textit{k}$ items of the Training Set as initial means (cluster centroids)
      \REPEAT
         \STATE $\textit{flag} \leftarrow false$ 
         \FOR{each item $t_{i}$ of the Training Set}
            \STATE Find the cluster $C$ which has the closest cluster centroid to $t_{i}$
            \IF{$C \neq$ current cluster of $t_{i}$} 
               \STATE Assign $t_{i}$ to $C$
               \STATE $flag \leftarrow true$ 
            \ENDIF
         \ENDFOR
         \STATE Compute new mean for each cluster
      \UNTIL{$flag == false$} \COMMENT{none item has moved to another cluster}      
      \FOR{each cluster $C$}
         \STATE $AvgDist_{C} \leftarrow$ Compute the the average distance of the items in $C$ from the Cluster Centroid
         \FOR{each item $t_{i}$ in $C$}
            \IF{Distance($t_{i}$, Centroid of \textit{C}) $\leq$ $D \times AvgDist_{C}$}
               \STATE Assign $t_{i}$  to the Core Set of \textit{C} ($CS_{C}$)
            \ELSE
               \STATE Assign $t_{i}$  to the Peripheral Set of $C$ ($PS_{C}$)
            \ENDIF
         \ENDFOR         
      \ENDFOR
   \STATE \COMMENT{Classification procedure}      
   \FOR{each unclassified item \textit{x}}
      \STATE identify \textit{L} nearest clusters (based on clusters centroids) from \textit{x}, $C_{1}, C_{2}, \ldots, C_{L}$ where $C_{1}$ is the nearest, $C_{2}$ is the second nearest and so on
      \IF {Distance($x$, Centroid of $C_{1}) \leq D \times AvgDist_{C}$}
          \STATE $R \leftarrow C_{1}$
      \ELSE    
          \STATE $R \leftarrow C_{1} \cup PS_{C_{2}} \cup PS_{C_{3}} \cup \ldots \cup PS_{C_{L}}$
      \ENDIF    
      \STATE Classify $x$ by executing the $k$-NN classifier over $R$
   \ENDFOR   
   
\end{algorithmic}
\end{algorithm*}

%\newpage

\section{REFERENCE SET REDUCTION THROUGH $k$-MEANS CLUSTERING}

The Reference Set Reduction method (for simplicity, RSRM) proposed by Hwang and Cho is an effective speed-up approach. The method is outlined in Algorithm~\ref{rsrm}. It uses the well-known $k$-means algorithm (McQueen, 1967) to find clusters in the training set (lines 2--13). Afterwords, each one cluster is divided into two sets which are called ``peripheral set" and ``core set". Particularly, the cluster items lying within a certain distance from the cluster centroid are placed into the ``core set", while the rest, more distant from the centroid, items are placed into the ``peripheral set" (lines 14--23).  

When a new item $x$ must be classified, the algorithm finds the nearest cluster $C_{1}$. If $x$ lies within the core area of $C_{1}$, it is classified by retrieving its $k$-nearest neighbors from $C_{1}$. Otherwise, the $k$ nearest neighbors are retrieved from the Reference Set $R$ formed by the items of the nearest cluster and the ``peripheral" items of the $L$ most adjacent clusters (lines 26--32). 

If the clusters were not divided and only the items of the nearest cluster were used to classify the new item (regardless of how distant from the centroid it was), many training items in the nearby clusters would be ignored. Thus, Hwang and Cho proposed the use of some nearby clusters as a safer approach. The main innovation in their method is that it uses only the peripheral items of these additional adjacent clusters. If all items (not only the peripheral) of these clusters were used, the computational cost would have been much higher.

A key factor of RSRM is the determination of the threshold that defines which items will be core and which peripheral. This is very critical since it determines how many items are accessed during classification. Hwang and Cho consider as peripheral items, those whose distance from the cluster centroid is greater than the double average distance among the items of each cluster. Thus, the average distance among the items in each cluster and the corresponding cluster centroid must be computed (line 15). 

\begin{table*}[th]
\caption{Dataset description (cost is in million distance computations)}
\centering 
\begin{tabular}{|c|c|c|c|c|c|c|c|c|}
\hline
\multirow{2}{*}{\textbf{dataset}} & \textbf{train/test} & \multirow{2}{*}{\textbf{attributes}} & \multirow{2}{*}{\textbf{classes}} & \textbf{best} & \textbf{accuracy} & \multirow{2}{*}{\textbf{cost}}\\
                 & \textbf{dataset size}      &       &                  & \textbf{k}   & (\%) &\\
\hline\hline
Letter recognition & 15000/5000 & 16 & 26 &4 & 95.68 & 75\\\hline
Magic gamma telescope& 14000/5020 & 10 & 2 & 12 & 81.39 & 70.28 \\\hline
Pendigits & 7494/3498 & 16 & 10 & 4 & 97.89 & 26.21 \\\hline
Landsat sattelite & 4435/2000 & 36 & 6 & 4 & 90.75 & 8.87 \\\hline
Shuttle & 43500/14500 & 9 &  7 & 2 & 99.88 & 630.75 \\\hline
\end{tabular}
\label{table:datasets}
\end{table*}

In this study, we do not use a particular threshold as Hwang and Cho did (they  used the double average distance). We introduce parameter $D$ to be responsible for the splitting of the clusters into core and peripheral sets. An item $x$ is placed into the peripheral set of cluster $C$, if: 
\begin{center}Distance($x$, centroid of $C$) $>$ $D \times AvgDist_{C}$\end{center}
For example, if $D$=1.5, the ``peripheral sets" include items that are more than 1.5 times the average distance away from the cluster centroid. The determination of $D$ is a critical issue and it should be made by considering the available number of clusters and the desirable trade-off between accuracy and computational cost.

Another issue that must be addressed is related to the number of clusters that are constructed (determination of the $k$ parameter in $k$-means algorithm) and the number of adjacent clusters that are examined when the new item lies within the peripheral area of the nearest cluster ($L$ parameter). Hwang and Cho empirically define $L=\lfloor\sqrt{k}\rfloor$.

\section{EXPERIMENTAL STUDY}

The extensive experimental study was conducted using five real life datasets distributed by the UCI Machine Learning Repository\footnote{\url{http://archive.ics.uci.edu/ml/}}. The datasets are presented in Table~\ref{table:datasets}. The fifth column lists the $k$ value found to achieve the highest accuracy when using the $k$-NN classifier over the whole training set (conv-$k$-NN). The computational cost was estimated by counting the distance computations needed to carry out the whole classification task. Of course, the cost measurements do not include the distance computations needed by the $k$-means clustering preprocessing procedure. Moreover, contrary to Hwang and Cho, who used the ROC distance metric in their experiment, we estimated all distances using the Euclidean distance. All datasets were used without data normalization or any other transformation. Also, in all RSRM experiments, we chose the $k$ values of the $k$-NN classifier that achieved highest accuracy (do not confuse this parameter with $k$ of $k$-means clustering).

We define $L=\lfloor\sqrt{k}\rfloor$ as Hwang and Cho did in their experiment. Concerning the $k$ parameter that determines the number of clusters that are formed, we built 8 classifiers for each dataset. Classifier$_{i}$ uses $k=\lfloor\sqrt{\frac{n}{2^{i}}}\rfloor$ clusters, $i$=1,\ldots,8, where $n$ is the number of items in the training set. Classifier$_{1}$ is based on the rule of thumb that defines $k=\lfloor\sqrt{\frac{n}{2}}\rfloor$ (Mardia et al, 1979). We decided to build classifiers that use low $k$ values based on the observation that Hwang and Cho set $k$=10 for a training set with 60919 items. For each classifier, we chose a varying value for $D$ (1, 1.5, and 2). Thus, we built and evaluated 8 * 3 = 24 classifiers for each dataset. 

\begin{figure}
\centering
\includegraphics[width=3cm]{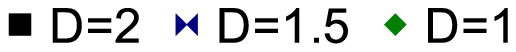}
\includegraphics[width=8cm]{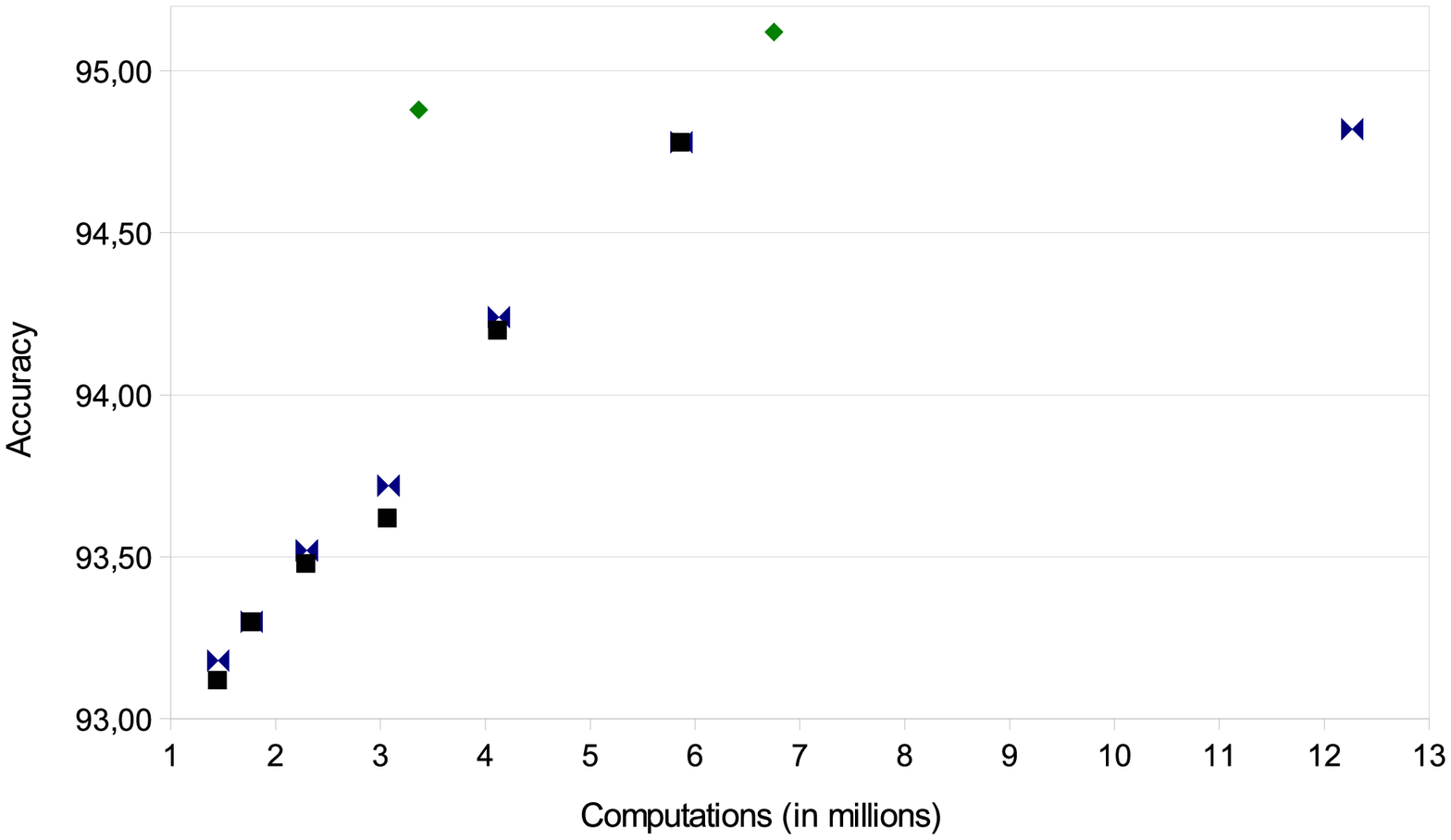}
\caption{Letter Image Recognition Dataset}
\label{fig:a1}
\vskip 1mm
\includegraphics[width=8cm]{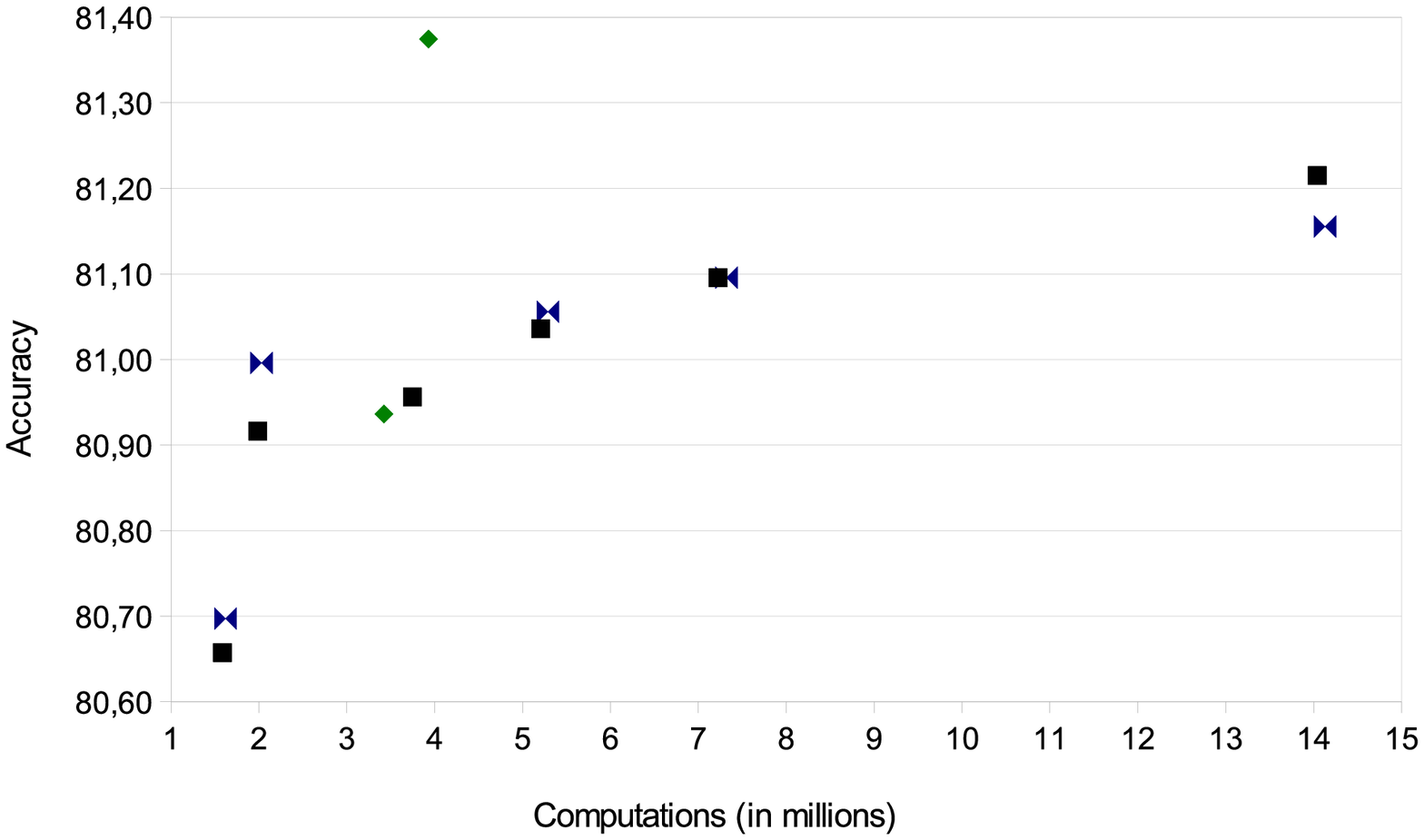}
\caption{Magic Gamma Telescope Dataset}
\label{fig:a2}
\end{figure}

\begin{figure}
\centering
\includegraphics[width=8cm]{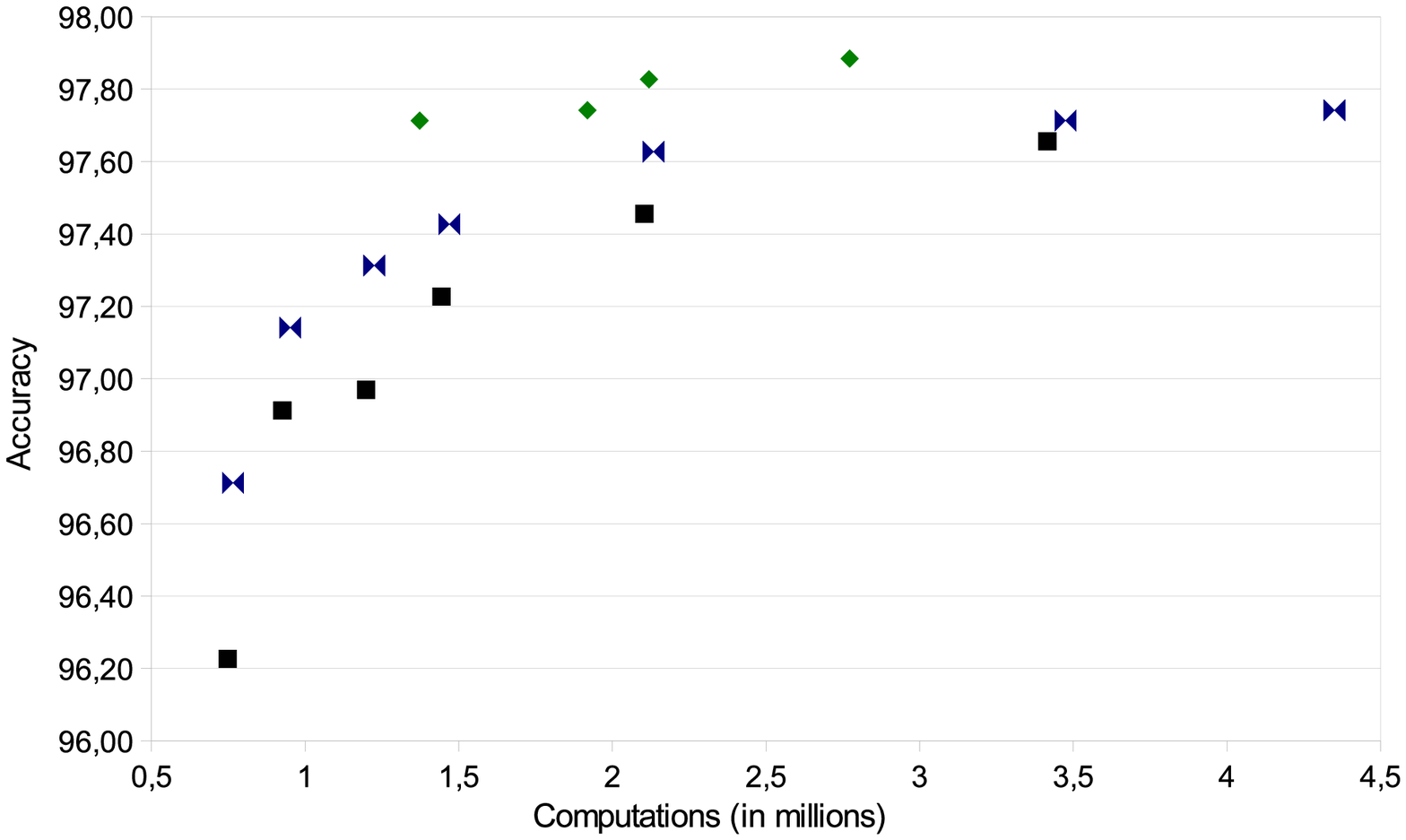}
\caption{Pendigits Dataset}
\label{fig:a3}
\vskip 1mm
\includegraphics[width=8cm]{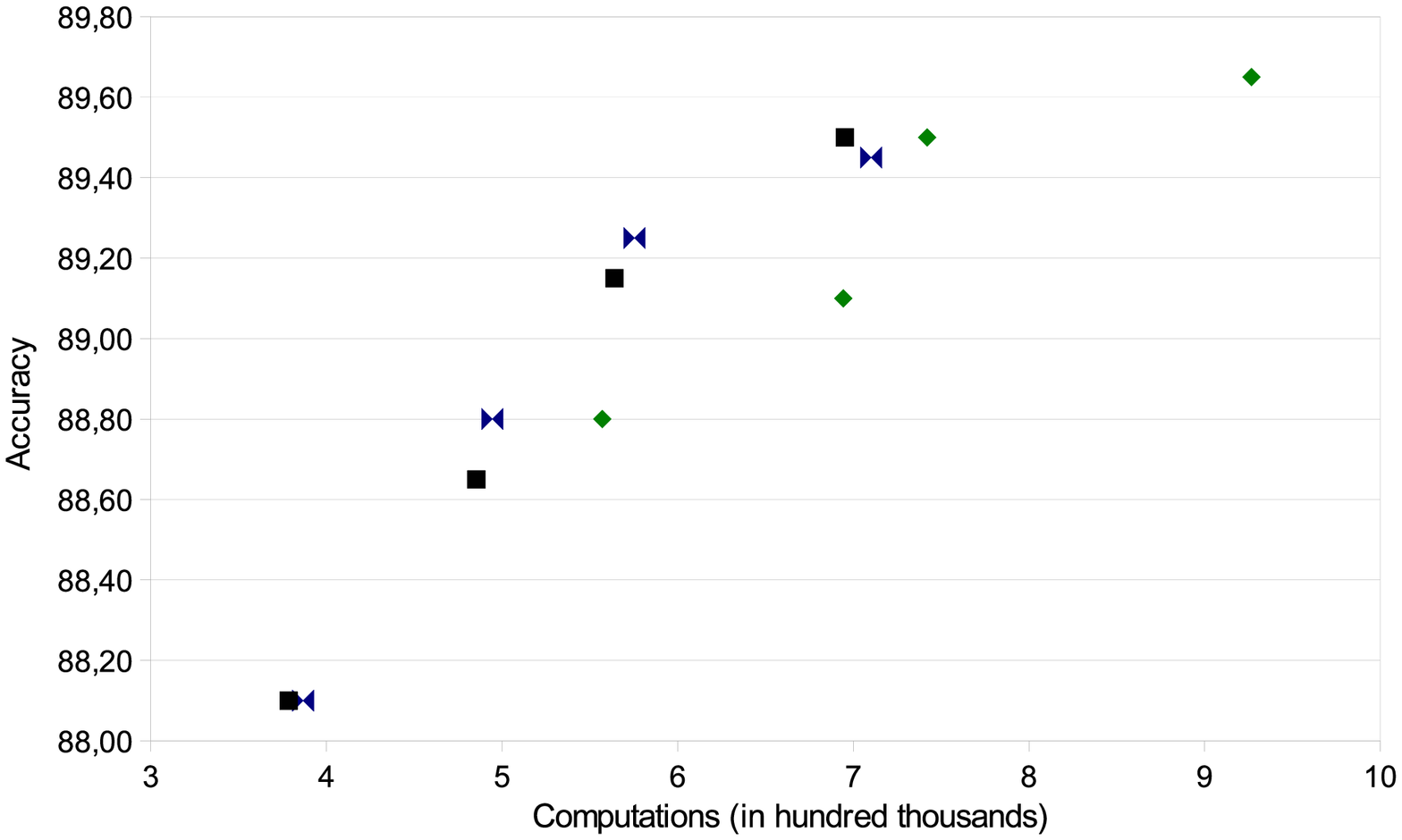}
\caption{Landsat Satellite Dataset}
\label{fig:a4}
\vskip 1mm
\includegraphics[width=8cm]{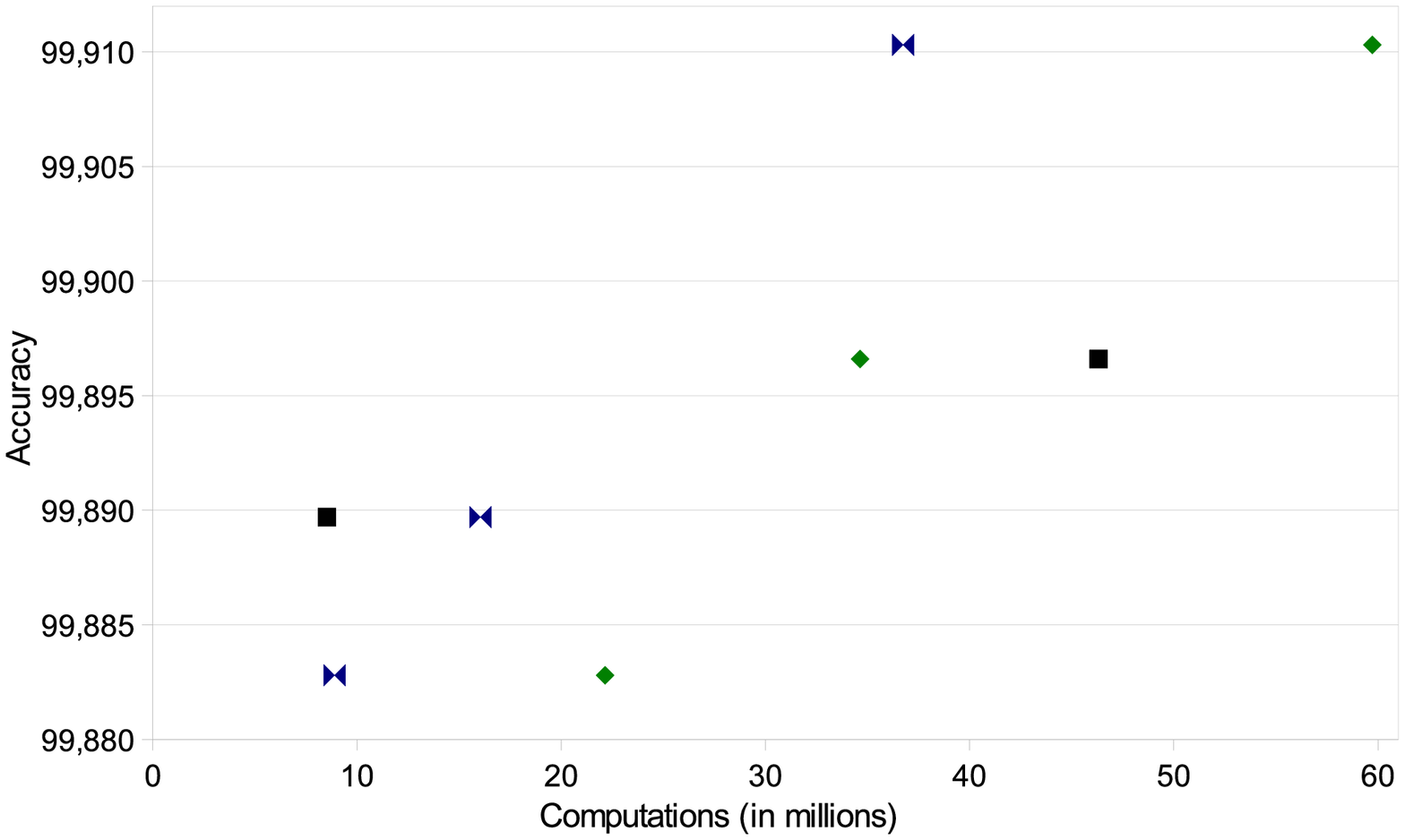}
\caption{Shuttle Dataset}
\label{fig:a5}
\end{figure}

In Fig.~\ref{fig:a1}--\ref{fig:a5}, for each dataset, the performance of the most accurate classifiers for a given cost are presented\footnote[2]{Detailed experimental results available at:\url{http://users.uom.gr/~stoug/RSRM.zip}}. The figures do not include the performance of conv-$k$-NN that is mentioned in Table~\ref{table:datasets}. In particular, in Fig.~\ref{fig:a1}--\ref{fig:a5}, the classifiers built by the three $D$ values (1, 1.5 and 2) are compared to each other. 

For the first three datasets (Fig.~\ref{fig:a1}--\ref{fig:a3}), the classifiers built for $D$=1 seem to perform better than the ones built for $D$=1.5 and $D$=2. In the cases of the Letter Image Recognition (LIR) and Magic Gamma Telescope (MGT) datasets, the superiority of the Classifiers$_{D=1}$ is obvious. In the case of LIR, the two Classifiers$_{D=1}$ presented in Fig~\ref{fig:a1} are build by setting $k$=$\lfloor\sqrt{\frac{15000}{2^{1}}}\rfloor$=86, $L$=$\lfloor\sqrt{86}\rfloor$=9 and $k$=$\lfloor\sqrt{\frac{15000}{2^{5}}}\rfloor$=21, $L$=$\lfloor\sqrt{21}\rfloor$=4, respectively. In MGT, the parameter values of the most accurate classifier are $D$=1, $k$=59 and $L$=7. Finally, in Pendigids, the fastest and slowest Classifier$_{D=1}$ is built by setting $k$=61 and $k$=15 respectively.

For the Landsat Satellite (LS) and Shuttle datasets (Fig.~\ref{fig:a4} -- \ref{fig:a5}) there is not a dominant $D$ parameter value in terms of performance and accuracy. In LS, the most accurate classifier is built by setting $D$=1 and $k$=16, while the fastest classifier that achieves an accuracy value over 89.2\% is built using $D$=1.5 and $k$=23. In Shuttle, the results are more confusing. This is because Shuttle is an imbalanced (skewed) dataset (approximately 80\% of the items belong to one class). However, in Shuttle, all classifiers presented in Fig.~\ref{fig:a5} manage to achieve higher accuracy than that of the conv-$k$-NN.

\section{CONCLUSION}

In this paper we presented an extensive experimental study on the Reference Set Reduction method through $k$-means clustering. In all experiments, the well-known Eucledian distance was used. The classification performance of RSRM depends on the determination of $k$ and $D$ parameters. In all cases, they should be adjusted by taking into consideration the application domain and the desirable trade-off between classification accuracy and computational cost. The experimental measurements indicate that if accuracy is more critical than cost, low $D$ and high $k$ and $L$ values (e.g. $D$=1) lead to an efficient classification method. On the other hand, if cost is more critical than accuracy, higher $D$ and lower $k$ and $L$ values may be more appropriate.

\end{document}